**A Deep Learning Model for Traffic Flow State Classification Based on Smart Phone Sensor Data**

**Wenwen Tu**
School of Transportation & Logistics
Southwest Jiaotong University
111, North 1st Section of Second Ring Road, Chengdu City, China
Department of Civil and Environmental Engineerin
University of Waterloo
Waterloo, N2L 3G1
Tel: (226)899-2946; Email: wenwen.tu@uwaterloo.ca

**Feng Xiao, Corresponding Author**
Big Data Research Center
Southwestern University of Finance and Economics
55, Guanghuacun Road, Chengdu City, China
Tel: 86-18200561506 Email: xiaofeng@swufe.edu.cn

**Liping Fu**
Department of Civil & Environmental Engineering
University of Waterloo
Waterloo, N2L 3G1
Tel. (519) 888-4567 ext. 33984 Email: lfu@uwaterloo.ca

**Guangyuan PAN**
Department of Civil and Environmental Engineerin
University of Waterloo
Waterloo, N2L 3G1
Email:p5pan@uwaterloo.ca

## ABSTRACT

This study proposes a Deep Belief Network model to classify traffic flow states. The model is capable of processing massive, high-density, and noise-contaminated data sets generated from smartphone sensors. The statistical features of Vehicle acceleration, angular acceleration, and GPS speed data, recorded by smartphone software, are analyzed, and then used as input for traffic flow state classification. Data from a five-day experiment is used to train and test the proposed model. A total of 747,856 sets of data are generated and used for both traffic flow states classification and sensitivity analysis of input variables. The result shows that the proposed Deep Belief Network model is superior to traditional machine learning methods in both classification performance and computational efficiency.

## INTRODUCTION

Timely and accurate monitoring of traffic congestion is the foundation of many Intelligent Transportation Systems. It allows government administration to make data-driven decisions and provides real-time traffic information for travelers, lowering system cost and enhance social efficiency. As technologies evolve, vehicle-based driving data can now be collected directly from smartphones. Collecting data from smartphones addresses many limitations of traditional fixed sensors, such as complex installations, high maintenance costs, and insufficient coverages. User penetration has been greatly improved as a result of ubiquitous use of smartphones. This paper is particularly concerned with applying data from smartphones as a means to estimate traffic states.

As discussed in Literature Review, a number of earlier efforts have attempted to infer traffic states based on data from Global Positioning System (GPS) traces. However, all of the approaches have relied on information of time-stamped speed data, which may contain large errors due to positioning deviations of the GPS units. For example, when traveling underground or in urban canyons, GPS signals may not be received accurately or may even be completely lost. As a result, GPS data could be incomplete or inaccurate. In addition, using speed solely may in some cases result in traffic analysis complication, such as misclassifications of transportation modes. In practice, misclassifications often happen to fast walking and slow biking, and to buses and cars on a congested road. Therefore, when speed information is not sufficient to distinguish different transportation modes or the GPS signal is lost, there is a need to develop methods that make use of additional data such as those from accelerometer for estimating traffic states.

Additionally, the massive and high-density data collected from smartphone sensor also contains noises. Hence, additional data processing and modeling are required to improve the accuracy of traffic flow state classification. However, traditional machine learning methods have been shown to have limitations in addressing massive noisy data. In contrast, deep learning model has achieved top-tier performance in many classification tasks in recent years. Yet, it has not been used in traffic flow state classification nor with the smartphone sensor data. Hence, the potential of the proposed model will be explored in this research.

This paper first provides a background review concerning the research of mobile devices data and deep learning networks in traffic analyses. Next, the proposed Deep Belief Network (DBN) model for traffic flow states classification is described in details. Smartphone sensor data was collected during a five-day experiment in Chengdu, China. In order to address the low accuracy and weak robustness issues from relying solely on speed, acceleration and angular acceleration data from smartphone sensor, combing GPS speed data, are used as inputs of traffic flow states classification. The classification results are then generated by the proposed DBN model. Finally, the proposed model's computational capabilities and robustness are discussed and compared with other machine learning models.

## LITERATURE REVIEW

Technology based on mobile devices has proven its usefulness in collecting activity-travel diary data. Mobile acquisition devices include GPS dedicated collection equipment, smartphone location and sensing devices (*1*). Comparing to conventional paper-based or phone-based data collection methods, data collection technology on mobile devices has been argued to reduce respondent and researcher burden. And the accuracy of the data would be better than those of conventional survey methods. Händel et al. analyzed approximately 4500 driving hours of road vehicle traffic data collected during the ten-month long project, i.e., the Berkeley Mobile Millennium Project. And then a framework was presented to deploy a smartphone-based measurement system for road vehicle traffic monitoring and usage-based insurance (*2*). Herrera presented a field experiment nicknamed Mobile Century, which was conceived as a proof of concept of a GPS-enabled cell phone based traffic monitoring system (*3*). Then they proposed and assessed methods to perform traffic state estimation in the presence of data provided by GPS-enabled cell phones. Furthermore, some researches also started to use GPS data obtained from smartphones to derive personal trip data (*4*). These researches either combined a web-based diary system or Geographic Information System (GIS), to receive additional information of transportation modes and trip purposes (*5-7*).

Furthermore, accelerometers have been used to identify the type of people's physical activity, such as walking, running, sitting and relaxing, watching TV, brushing teeth, and climbing (*8*). Recent research has attempted to combine GPS and accelerometer data to recognize transportation modes (*9, 10*). For example, Cooper et al. combined accelerometer and GPS data to investigate the transportation modes of children uses to attend school (*11*). Moreover, a few studies have also attempted to detect transportation modes using accelerometer data from smartphone sensors (*12*). Researchers also find that the smartphone-based algorithms can accurately detect four distinct patterns (braking, acceleration, left cornering and right cornering) (*13*). Drivers' aggressive and risky behavior can be captured by the integrated GPS and accelerometer sensors based on smartphone (*14*). However, accelerometers record accelerations in three dimensions, which do not directly reflect the differences in transportation modes. Therefore, enhanced algorithms are required to better differentiate between different transportation modes, traffic network congestion classification (*9, 15*).

In the past decades, many traditional machine learning methods were used in the traffic states classification. However, with a great amount of noisy data from mobile sensor, computing time and calculation accuracy were far from ideal (*16*). The challenge of predicting traffic flows are the sharp nonlinearities due to transitions between free flow, breakdown, recovery, and congestion. In recent years, Deep Neural Network (DNN) has become a great success in processing massive data. DNN is a type of artificial neural networks (ANN), so it obeys the universal approximation theory. This ensures a neural network has global approximate ability for any nonlinear function if enough hidden units are given. Furthermore, by using a deep structure, DNN overcomes the shortcomings of exponential explosion and insufficient feature learning of traditional ANN such as Multi-layer Perceptron (MLP) (*17, 18*). Some researchers have explored this approach in several transportation tasks like traffic flow prediction and traffic sign classification. A deep architecture model was applied using autoencoders, and the model built blocks to represent traffic flow features for prediction. The experiments prove that the proposed model for traffic flow prediction has good performance (*19*). Huang et al. proposed a deep architecture for traffic flow prediction. The model consists of two parts, a deep belief network (DBN) at the bottom and a multi-task regression layer at the top. The results from experiments of traffic flow data sets demonstrate its very good performance to learn traffic flow features (*20*). Deep learning architectures are proven to capture these nonlinear spatio-temporal effectively (*21*). So in this research, we decided use a deep belief network (DBN) as an alternative for traffic state classification.

## THE MODEL

An illustration of the DBN model is shown in between units of the same layer. In the training process, each hidden layer extracts the last layer's data features to form a better, although more abstract, distributed representation of input data. The last layer is the output layer that has the exact units associated with the classes.

FIGURE **1**. It consists of three types of layers. The first is the visible layer that receives the original feature data and acts as the input layer. This input layer is followed by three hidden layers, with 300 neurons in each layer. Input layer and hidden layer 1 forms the first Restricted Boltzmann Machines (RBM), and hidden layer 1 and hidden layer 2 forms the second RBM, and so on. The structure of each RBM is of a two-way full connectivity between two layers. No connection exists

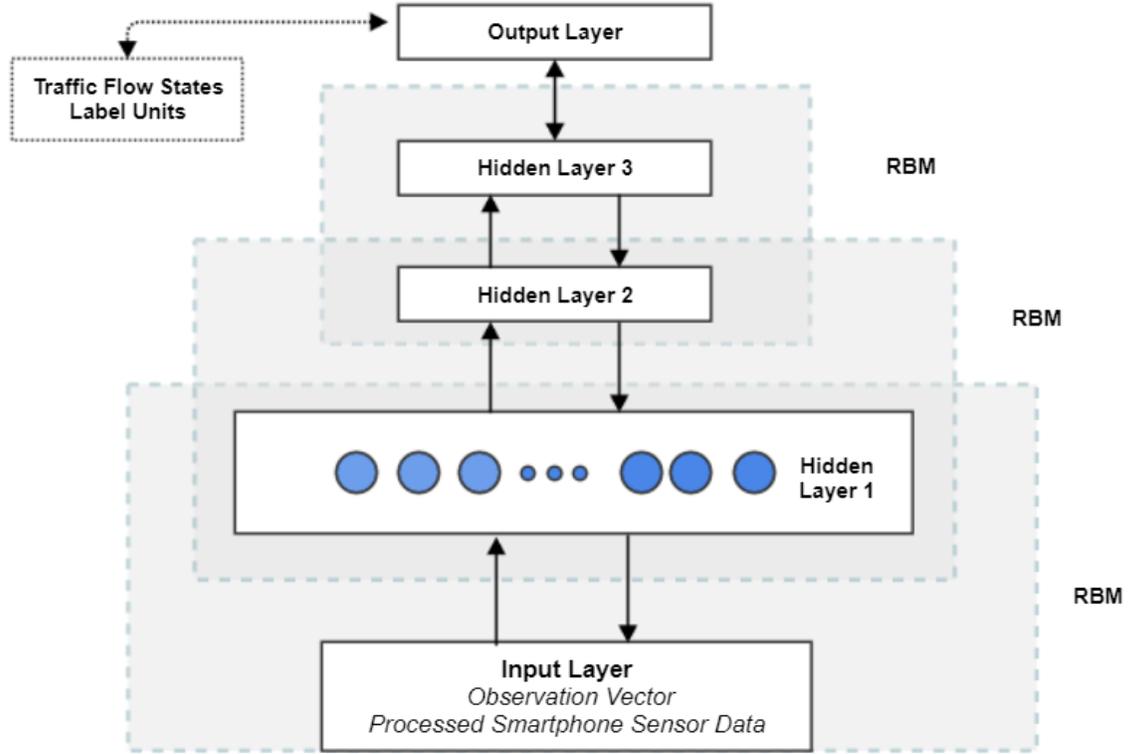

**FIGURE 1 The structure of the DBN traffic flow states classification model**

RBM can be stacked and trained in a greedy manner to form Deep Belief Networks (DBN). DBN are graphical models which learn to obtain a deep hierarchical training data representation. Typically, all visible units are connected to all hidden units. The weights on the connections and the biases of the individual units define a probability distribution over the binary state vectors v of the visible units via an energy function. The energy of the joint configuration $(\mathbf{v}, \mathbf{h})$ is given by Eq. (1).

$$E(\mathbf{v}, \mathbf{h}; \theta) = -\sum_{i=1}^{V} \sum_{j=1}^{H} w_{ij} v_i h_j - \sum_{i=1}^{V} b_i v_i - \sum_{j=1}^{H} a_j h_j \qquad (1)$$

Where $\theta = (w, b, a)$, $w_{ij}$ represents the symmetric interaction term between visible unit $i$ and hidden unit $j$ while $b_i$ and $a_j$ are their bias terms. $V$ and $H$ are the numbers of visible and hidden units. The probability that the model assigns to a visible vector $\mathbf{v}$ is given in Eq. (2).

$$p(\mathbf{v}; \theta) = \frac{\sum_{\mathbf{h}} e^{-E(\mathbf{v}, \mathbf{h})}}{\sum_{\mathbf{u}} \sum_{\mathbf{h}} e^{-E(\mathbf{u}, \mathbf{h})}} \qquad (2)$$

Since there are no hidden-hidden or visible-visible connections, the conditional distributions $p(\mathbf{v}\,|\,\mathbf{h})$ and $p(\mathbf{h}\,|\,\mathbf{v})$ are factorial and are given by Eqs. (*3*) and (*4*).

$$p(h_j = 1\,|\,\mathbf{v};\theta) = \sigma(\sum_{i=1}^{V} w_{ij}v_i + a_j) \qquad (3)$$

$$p(v_j = 1\,|\,\mathbf{h};\theta) = \sigma(\sum_{j=1}^{H} w_{ij}h_i + b_j) \qquad (4)$$

Where $\sigma(x) = (1 + e^{-x})^{-1}$. RBMs have an efficient approximate training procedure called "contrastive divergence" which makes them suitable as building blocks for learning DBN. We repeatedly update each weight, $w_{ij}$, using the difference between two measured, pairwise correlations:

$$\Delta w_{ij} = \left\langle v_i h_j \right\rangle_{data} - \left\langle v_i h_j \right\rangle_{model} \qquad (5)$$

The first term in Eq. 错误!未找到引用源。 is the measured frequency with which visible unit $i$ and hidden unit $j$ are on together when the visible vectors are samples from the training set and the states of the hidden units are determined by Eq. 错误!未找到引用源。. The second term is the measured frequency with which $i$ and $j$ are both on when the visible vectors are "reconstructions" of the data vectors and the states of the hidden units are determined by applying Eq. 错误!未找到引用源。 to the reconstructions. Reconstructions are produced by applying Eq. 错误!未找到引用源。 to the hidden states that are computed from the data when computing the first term of the Eq. 错误!未找到引用源。.

After unsupervised pre-training, a set of labels will be attached to the top to present the classification categories. Then, traditional back propagation (BP) is applied to fine-tune the model. The DBN model is better than BP because BP is merely training a randomly initialized network. The part of BP algorithm in DBN model only needs a local search in the parameter weight space. So comparing to a global space search, the local search is faster with higher accuracy.

## EXPERIMENTAL SETUP AND RESULTS
### Data collection
In the data collection experiment, the hardware is a smartphone with motion sensors, placed in a fixed position in a moving vehicle. A smartphone app is used for data acquisition. The data acquisition module utilizes the hardware motion sensors, such as accelerometers, triaxle gyroscopes, and GPS modules to detect real-time motion data such as speed, acceleration and angular acceleration.

In this paper, the 2nd Ring Road in Chengdu is selected as the experimental site due to its access controlled status and good road conditions. The current traffic state is obtained from Variable Message Signs (VMS) based on video data. Traffic congestion status, provided by software GAODE map, is based on GPS floating car data. The results of comparing the above two statuses, obtained at the same time, were found to be relatively consistent. Therefore, this traffic state data was considered reliable to be the output labels for training the classification model. For the sake of simplicity, traffic flow is divided into three states: free flow, steady flow, and congested flow. There are eight variables in this experiment: acceleration in three axis directions AX, AY, AZ; angular acceleration in three axis directions GX, GY, GZ; speed; and corresponding traffic flow states.

A total of 747,856 observations were obtained from this five-day long experiment, as shown in

FIGURE **2**. Each observation contains eight variables of AX, AY, AZ, GX, GY, GZ, speed, and corresponding traffic flow states. 45.15% of the data is in the free flow state, 30.04% of the data is in the steady flow state, and the remaining 24.81% is in the congested flow. The distribution of the data in the three traffic flow states can be considered reasonably balanced.

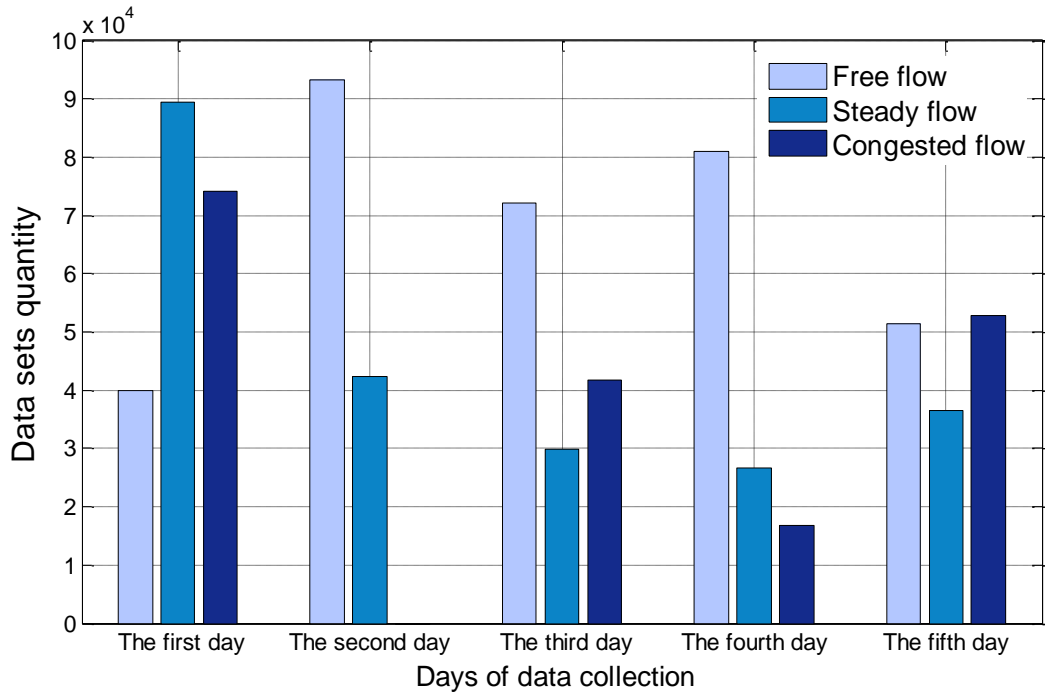

**FIGURE 2 The collected data with corresponding traffic flow states.**

The accelerometer in the smartphone is a spring damped oscillator. Its working principle is that the mass block moves in the opposite direction due to inertial force when the smartphone accelerates. A proportional voltage signal is then shown as a result of that. The voltage signal represents the magnitude of the smartphone acceleration. Acceleration data is divided into three variables of AX, AY, and AZ, representing the acceleration value respectively in x-axis, y-axis, and z-axis three directions (FIGURE 3).

The built-in three-axis gyroscope, called a micromechanical gyroscope, is used to measure the angle of change and the direction to be maintained. According to the law of conservation of angular momentum, the angular acceleration of the rotational axis indirectly reflects the external force. GX, GY, and GZ represent the angular acceleration in the three directions in x-axis, y-axis, and z-axis respectively.

The GPS module is integrated in the mobile phone board radio frequency chip module. It can track the trajectory, and also provide time and space with the latitude and longitude coordinates. Speed can then be calculated accordingly.

The software retrieves the data from the accelerometer and the triaxle gyro module at a frequency of 50 Hz. Each data contains time, sensor call records, accelerometers, and three-axis data, which are all collected by the three-axis gyroscope. In the meantime, the software collects the latitude and longitude of current location, along with current speed of movement every second.

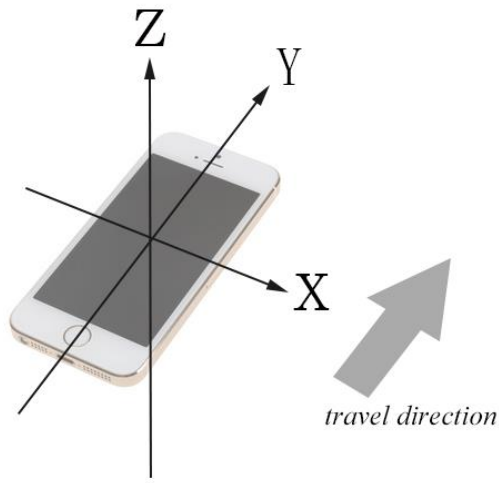

**FIGURE 3 Three axis coordinate directions of the three-axis gyroscope.**

**Data pre-processing**
The acceleration data (AX, AY, AZ, GX, GY, GZ) and speed data are arranged in chronological order, as shown in FIGURE 4, where light blue indicates that the current traffic is free flow; blue indicates the current traffic is steady flow; and black indicates the current traffic state is congested flow. Comparing to congested flow, data has greater amplitude in free flow as a result of different vehicle speed, i.e, vehicle in free flow travels faster with a more constant speed. The data also shares a similarity with voice signal, as it fluctuates around its mean value.

AX, AY, and AZ represent the amount of acceleration in Gs (where G is one unit of gravity) for that axis respectively. For example, if the smartphone is stationary and placed vertically in portrait orientation, it would have acceleration of (AX: 0, AY: -1, AZ: 0); laying flat on its back on a surface would be (AX: 0, AY: 0, AZ: -1). The smartphone, used in this experiment, is placed on its back on a flat surface in the car.

After combining the calculated boxplot of AX, AY, AZ, GX, GY, GZ, and speed variables (FIGURE 5), most of data are found to be concentrated at and fluctuates around the value of 0. Only variable AZ is concentrated at value of -1. Variable AZ also contains more data anomalies and greater fluctuations.

According to the boxplot 错误!未找到引用源。, distributions of acceleration and angular acceleration data are alike in all three states. If the data cannot show the statistical features of the moving vehicle, further statistical analysis on the data is required, to ensure the traffic flow state classification accuracy.

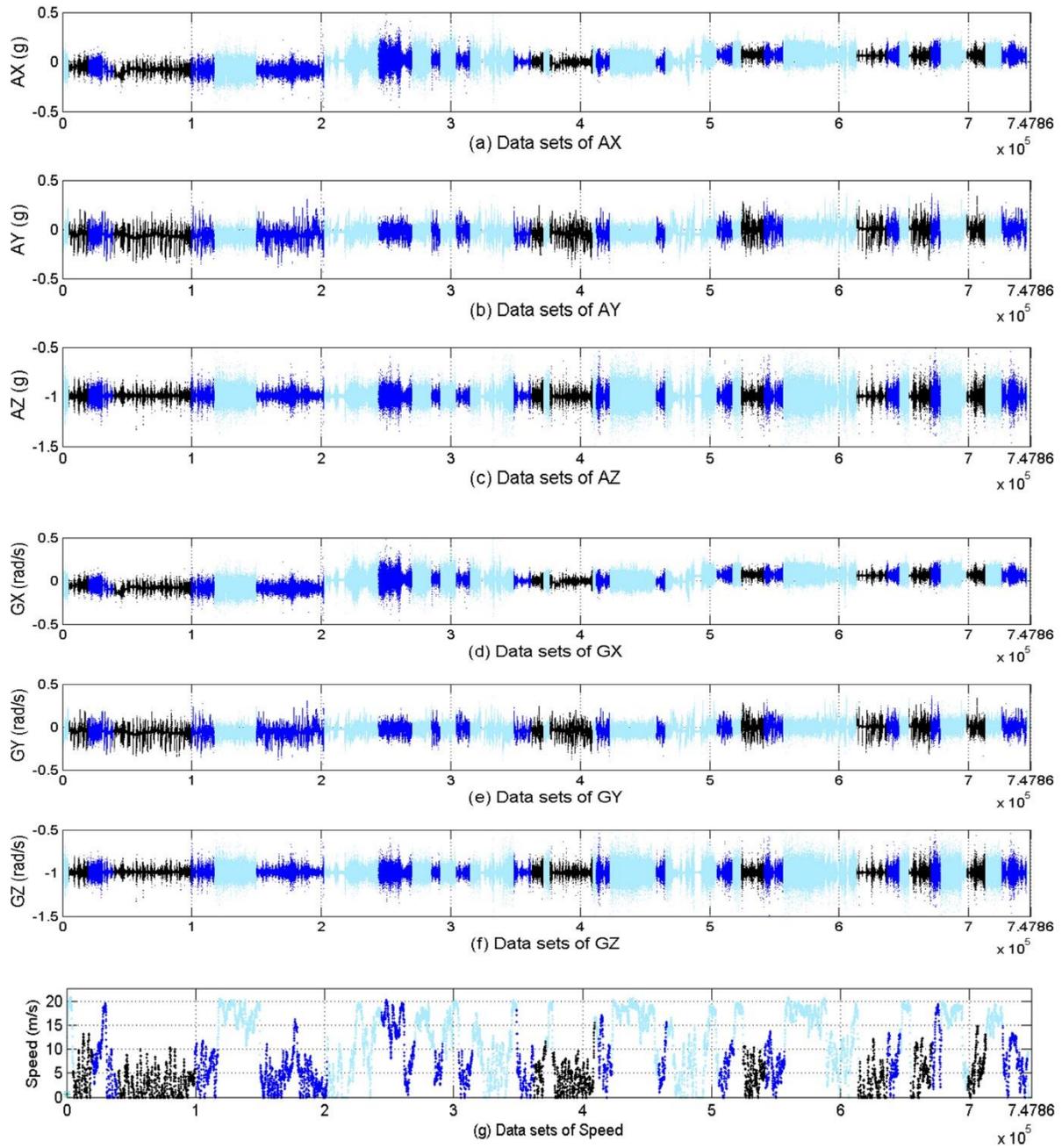

**FIGURE 4 The time series data of AX, AY, AZ, GX, GY, GZ, and speed**

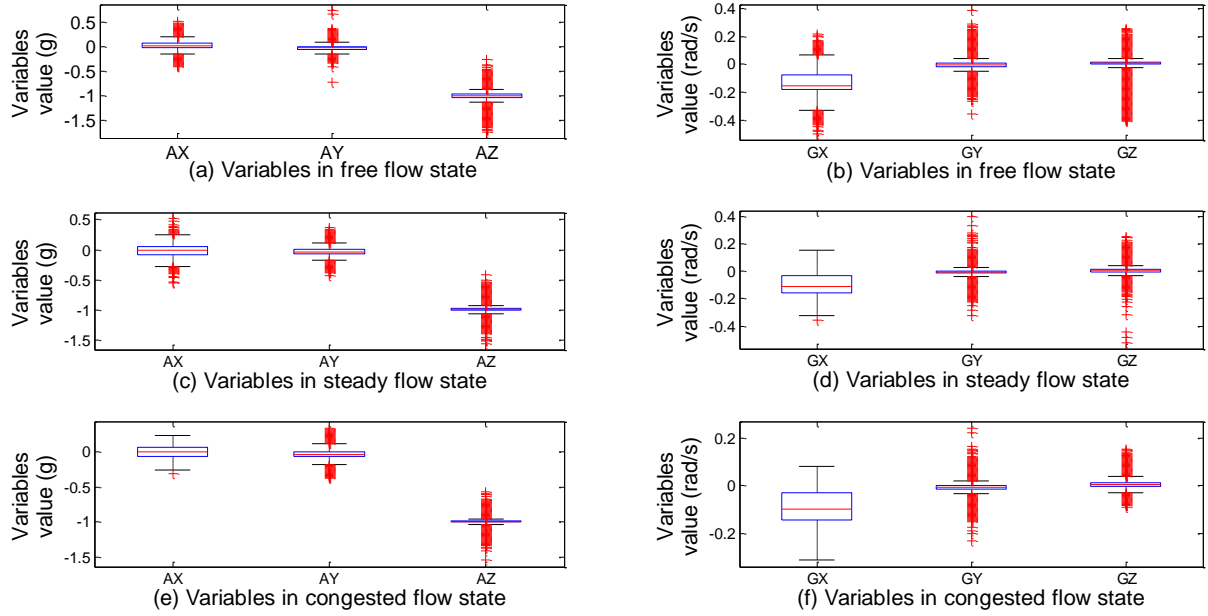

**FIGURE 5 Boxplot of AX, AY, AZ, GX, GY, and GZ data in different traffic flow states.**

A two-step feature calculation process is followed to extract the hidden information behind each variable data. The first step is to use data from a moving window of $n_1$ data points to calculate various new statistical features such as range, standard deviation, mean, quartile, variance, average absolute deviation, skewness, kurtosis, and coefficient of variation. The calculation process moves forward at a step size of $m_1$ data points and is repeated until it reaches the end of data stream. In the second step, in order to further extract the data features in different states, we calculate the number of these statistical eigenvalues that are greater than a certain threshold based on the result from the first step. The threshold is taken accordingly from TABLE 1. The methodology of step two is the same as the first step, $n_2$ data points are used each time for calculation and it moves forwards at a step size of $m_2$ data points, until all of the data are processed.

**TABLE 1 The selected statistical features and thresholds**

| Variable Number | Statistical Feature | Variable | Threshold | Unit |
|---|---|---|---|---|
| 1 | Standard Deviation | AX | 0.41 | g |
| 2 | Mean | speed | 0.9 | m/s |
| 3 | Quartile | AX | 0.41 | g |
| 4 | Quartile | AZ | 0.51 | g |
| 5 | Variance | AZ | 0.2 | $g^2$ |
| 6 | Average Absolute Deviation | AX | 0.4 | g |
| 7 | Average Absolute Deviation | GX | 0.25 | rad/s |
| 8 | Coefficient of Variation | AZ | 0.5 | / |
| 9 | Range | AX | 0.2 | g |
| 10 | Range | AZ | 0.2 | g |
| 11 | Standard Deviation | AX | 0.3 | g |
| 12 | Standard Deviation | AZ | 0.2 | g |
| 13 | Standard Deviation | GX | 0.1 | rad/s |
| 14 | Standard Deviation | GY | 0.15 | rad/s |
| 15 | Mean | speed | 0.4 | m/s |
| 16 | Quartile | AX | 0.18 | g |
| 17 | Quartile | AZ | 0.26 | g |
| 18 | Variance | AX | 0.08 | $g^2$ |
| 19 | Variance | AZ | 0.06 | $g^2$ |
| 20 | Average Absolute Deviation | AX | 0.26 | g |
| 21 | Average Absolute Deviation | AZ | 0.25 | g |
| 22 | Average Absolute Deviation | GX | 0.2 | rad/s |
| 23 | Coefficient of Variation | AZ | 0.22 | / |

After the two-step feature calculation and data normalization, we can find that the data in different states are obviously separated from each other, as shown in

FIGURE **6**. The data label on boxplot's x-axis corresponds to the variable number from TABLE 1. Comparing with the results from the first step, the mean values gradually decrease from free flow, to steady flow, and to congested flow. However, there is still a lot of noisy data and outliers, implying that it requires the classification model to be robust in dealing with noisy data.

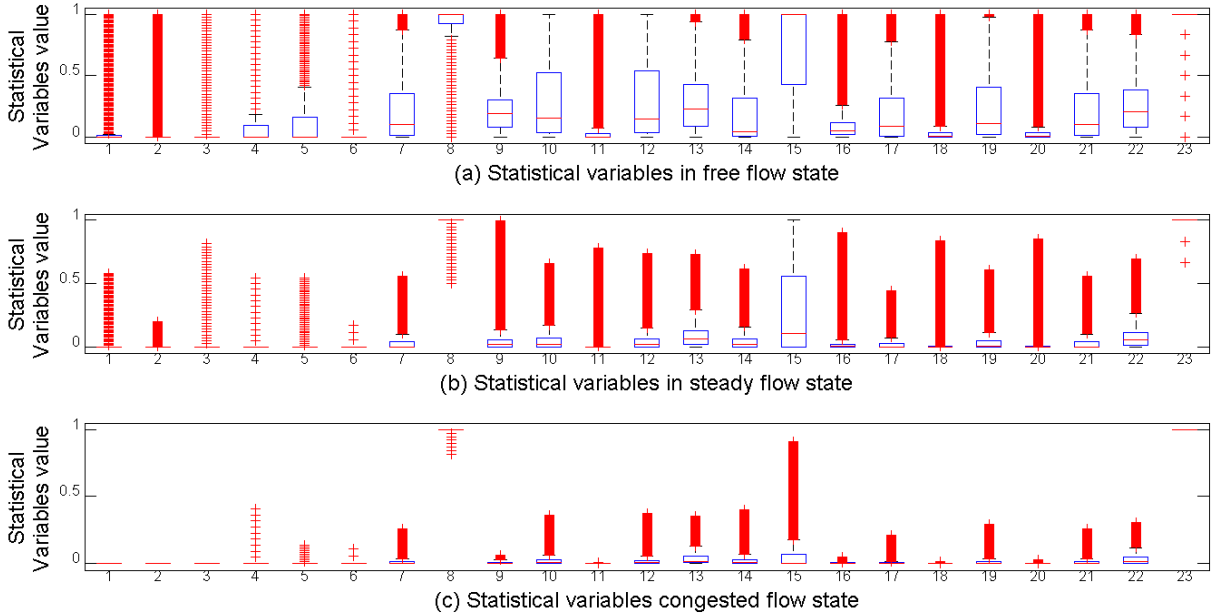

(a) Statistical variables in free flow state

(b) Statistical variables in steady flow state

(c) Statistical variables congested flow state

**FIGURE 6 Boxplot of statistical variables in different traffic flow states.**

## Analysis of classification results

We randomly divide the data into ten sets, and selecte seven sets as the training data and the remaining for testing. The above process is repeated ten times. A fixed value of 30 is set for the unsupervised training iterations in the DBN model. Unsupervised training learning rate is set at 2. Supervised training iteration value is changed accordingly to evaluate the classification errors, as shown in

FIGURE **7**. The average error of the ten test groups decreases from 9.91% to 3.93% as the supervised training iteration increases from 20 to 200. The average error drops to 2.44% when the iteration reaches 1000. Based on the result, the proposed DBN traffic flow states model demonstrates a powerful learning capability.

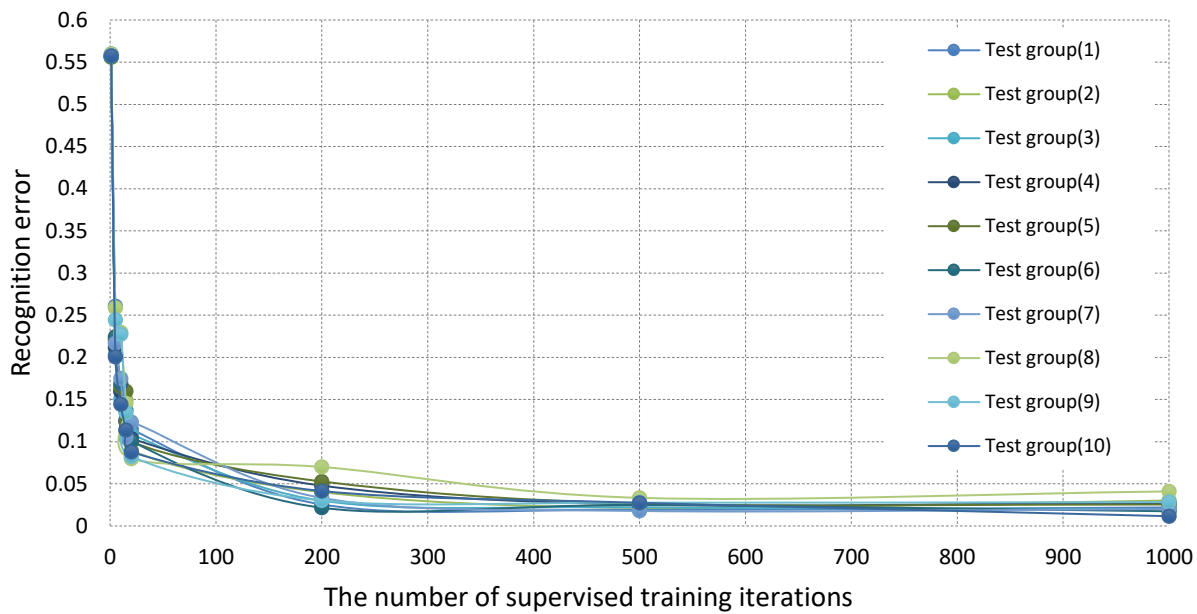

**FIGURE 7 Classification error of DBN model with ten test groups.**

As a comparative experiment, we use Support Vector Machine (SVM) method, Discriminant Analysis Classifier method, the Ensemble method with adaptive boosting and decision Tree, and Naive Bayes Classifier algorithm for modeling references. Finally, we also consider the situation that uses only GPS speed data as input variables. Accuracy test of each group are calculated using DBN model, the results are shown in

FIGURE **8**:

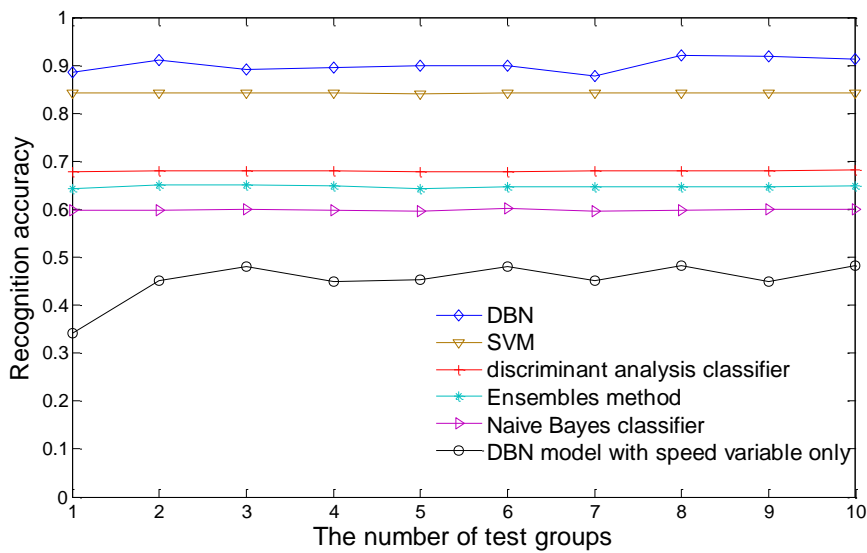

**FIGURE 8 Classification accuracy with DBN model and other machine learning models**

The average accuracy of test groups with DBN model is at 90.09%, and its average computing time is 519.734 s. The average accuracy of SVM model is at 84.22%, and its average calculation time is 7040.712 s. DBN model's average calculation time is only 7.38% of SVM model's, but the average accuracy is higher by 6.97%. We use some other traditional machine learning algorithms to calculate the same data set, and the result is that the discriminant analysis classifier method has an average accuracy of 68.00%, the Ensembles method 64.71%, and Naive Bayes Classifier method 59.80%. In contrast, the average accuracy of the DBN model increases the accuracy of the three machine learning methods by 32.48% and 39.21% and 50.65%, respectively. As a comparative experiment, we use only the speed variable as input and use DBN model to calculate the classification. Its average accuracy is only at 45.25%. Therefore, using only smartphone GPS speed data is insufficient to achieve highly accurate classification results. It demonstrates that acceleration and angular acceleration variables can provide additional information to describe the traffic flow states, and ultimately improves the state classification significantly.

## Sensitivity analysis

To evaluate the impact of parameters $m_1$, $n_1$, $m_2$, and $n_2$ on classification accuracy of traffic flow states, we use different values of the four parameters to calculate the classification accuracy of DBN model. The results are shown in FIGURE **9**. In the first data feature calculation, the proposed classification method is not sensitive to the parameters $m_1$ and $n_1$. Despite a large range of changes of $m_1$ and $n_1$, the accuracy of the ten experiments remain high. But when parameter $n_1$ reaches a small value, such as 10, some of the test groups showed slightly reduced accuracy. Nevertheless, the accuracy of most groups remains above 88%.

In addition, in the second feature calculation, the model accuracy becomes more sensitive to parameters $m_2$ and $n_2$ with the classification accuracy decreasing as $m_2$ increases and $n_2$ decreases. The reason is that the second calculation has less data but contains more feature information compared to the first calculation. When the number of moving sets is very large or the amount of combined data is very small, feature learning is more likely to be affected due to loss of information. However, in general, the deep learning-based traffic state classification method has stable performance as long as each of the parameters $m_1$, $n_1$, $m_2$, and $n_2$ is assigned with a value within a reasonable range.

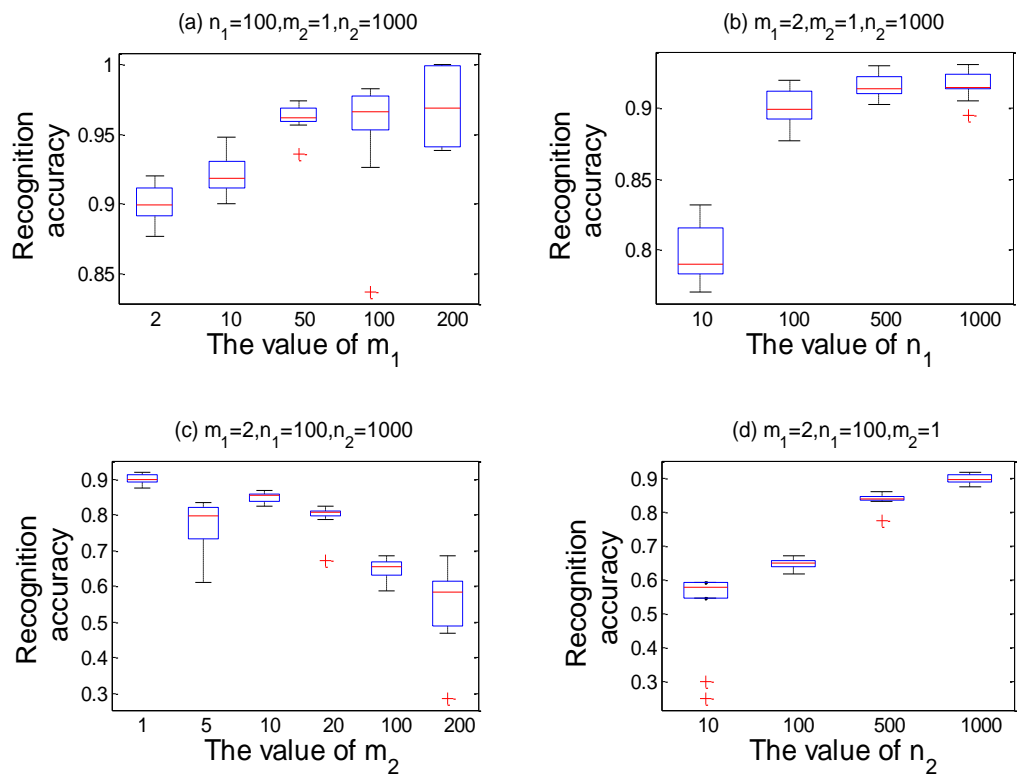

**FIGURE 9 Classification accuracy with different parameters** $m_1$, $n_1$, $m_2$, **and** $n_2$

## CONCLUSIONS AND OUTLOOK

In this paper, we propose a new deep learning based model - Deep Belief Network for automatic classification of traffic flow states based on data from a smartphone. The smartphone data includes time stamped acceleration and angular acceleration rates and GPS speed readings. The former two are introduced to address the low accuracy problem caused by solely relying on GPS speed data. The DBN model is trained and tested using real data collected from a smartphone application and shown to perform exceptionally well in comparison to the traditional machine learning models.

To evaluate the proposed model, we conduct a five-day experiment, yielding a total of 747,856 observations. We find that acceleration data has small variations under different traffic flow states, with small changes in its mean. The same pattern is also found from angular acceleration data. Therefore, to extract the hidden information, we propose additional features that can be derived from the original data, such as range, standard deviation, mean, quartile, and variance. Each feature values above a certain threshold are used as the input variables for traffic flow state classification. The new features along with the original data are used as input to the DBN model. As the number of top level supervised training iterations increases, the classification accuracy of the 10 test groups inceases and finally converges at 97.57%. The highest accuracy recorded is 98.26%. In contrast, using the same data sets, the SVM, discriminant analysis classifier method, the ensemble method with adaptive boosting and decision Tree, and Naive Bayes classifier algorithm, yield only accuracies of 84.22%, 68.00%, 64.71% and 59.80%, respectively. The result demonstrates that Deep Belief Network is capable of achieving higher classification accuracy in a shorter period of training time compared to traditional machine learning methods when the data set is relatively big and has a relatively high dimension.

As a point of outlook, traffic flow states and driver behaviors could also be investigated from smartphone sensor data such as GPS speed, acceleration, and angular acceleration. Such a method can be valuable for expanding the types of the data that are usually collected by conventional traffic data collection systems. As a result, both travelers and government agencies can gain better information about traffic patterns and conditions. Nevertheless, further research should be conducted to test the performance of the deep learning model under a wider range of conditions with larger datasets.

## Acknowledgements

The research is supported by the National Natural Science Foundation of China (71622007, 71431003) and a grant from the Fundamental Research Funds for the Central Universities (JBK170501).

# REFERENCES

1.  Bao, L., Intille, S.S., *Activity Recognition from User-Annotated Acceleration Data, International Conference on Pervasive Computing*. Springer, 2004.

2.  Byon, Y.J., Abdulhai, B., Shalaby, A.S., Impact Of Sampling Rate Of GPS-Enabled Cell Phones on Mode Detection and GIS Map Matching Performance, *Transportation Research Board 86th Annual Meeting*, No. 07-1795, 2007, pp. 21.

3.  Cooper, A.R., Page, A.S., Wheeler, B.W., Griew, P., Davis, L., Hillsdon, M., Jago, R., Mapping the Walk to School Using Accelerometry Combined with a Global Positioning System. *American Journal of Preventive Medicine*, Vol.38, No. 4, 2010, pp. 178-183.

4.  Händel, P., Ohlsson, J., Ohlsson, M., Skog, I., Nygren, E., Smartphone-Based Measurement Systems for Road Vehicle Traffic Monitoring and Usage-Based Insurance. *IEEE Systems Journal*, Vol.8, No.4, 2014, pp. 1238-1248.

5.  Herrera, J.C., *Assessment of GPS-Enabled Smartphone Data And Its Use in Traffic State Estimation for Highways*, University of California, Berkeley, 2009.

6.  Hinton, G.E., Sejnowski, T.J., Learning and Releaming in Boltzmann Machines. *Parallel Distrilmted Processing*, 1986.

7.  Huang, W., Song, G., Hong, H., Xie, K., Deep Architecture for Traffic Flow Prediction: Deep Belief Networks with Multitask Learning. *IEEE Transactions on Intelligent Transportation Systems*, Vol.15, No.5, 2014, pp. 2191-2201.

8.  Lecun, Y., Bengio, Y., Hinton, G., Deep Learning. *Nature*, Vol.521, No.7553, 2015. pp. 436-444.

9.  Lee, Y.S., Cho, S.B., Activity Recognition with Android Phone Using Mixture-of-Experts Co-Trained with Labeled and Unlabeled Data. *Neurocomputing*, Vol.126, 2014, pp. 106-115.

10. Lv, Y., Duan, Y., Kang, W., Li, Z., Wang, F.Y., 2015. Traffic Flow Prediction With Big Data: A Deep Learning Approach. *IEEE Transactions on Intelligent Transportation Systems*, Vol.16, No.2, pp. 865-873.

11. Oliver, M., Badland, H., Mavoa, S., Duncan, M.J., Duncan, S., Combining GPS, GIS, and Accelerometry: Methodological Issues in the Assessment of Location and Intensity of Travel Behaviors. *Journal of Physical Activity and Health*, Vol.7, No.1, 2010, pp. 102-108.

12. Pereira, F., Carrion, C., Zhao, F., Cottrill, C.D., Zegras, C., Ben-Akiva, M., the Future Mobility Survey: Overview and Preliminary Evaluation, *Proceedings of the Eastern Asia Society for Transportation Studies*. Vol.9, 2013.

13. Polson, N.G., Sokolov, V.O., Deep Learning for Short-Term Traffic Flow Prediction. *Transportation Research Part C: Emerging Technologies*, Vol.79, 2017, pp. 1-17.

14. Predic, B., Stojanovic, D., Enhancing Driver Situational Awareness through Crowd Intelligence. *Expert Systems with Applications*, Vol.42, No.11, 2015, pp. 4892-4909.

15. Reddy, S., Burke, J., Estrin, D., Hansen, M., Srivastava, M., Determining Transportation Mode on Mobile Phones, *2008 12th IEEE International Symposium on Wearable Computers,* 2008, pp. 25-28.

16. Schmidhuber, J., Deep Learning in Neural Networks: An Overview. *Neural Networks* Vol.61, 2015, pp. 85-117.

17. Troped, P.J., Oliveira, M.S., Matthews, C.E., Cromley, E.K., Melly, S.J., Craig, B.A., Prediction of Activity Mode with Global Positioning System and Accelerometer Data. *Medicine and Science in Sports and Exercise*, Vol.40, No.5, 2008, pp. 972-978.

18. Vlahogianni, E.I., Barmpounakis, E.N., Driving Analytics Using Smartphones: Algorithms, Comparisons and Challenges. *Transportation Research Part C: Emerging Technologies*, Vol.79, 2017, pp. 196-206.

19. Xu, D., Song, G., Gao, P., Cao, R., Nie, X., Xie, K., Transportation Modes Identification from Mobile Phone Data Using Probabilistic Models, *Advanced Data Mining and Applications,* 2011, pp. 359-371.

20. Zhang, L., Dalyot, S., Eggert, D., Sester, M, Multi-Stage Approach to Travel-Mode Segmentation and Classification of GPS Traces, *International Archives of the Photogrammetry, Remote Sensing and Spatial Information Sciences:[Geospatial Data Infrastructure: From Data Acquisition And Updating To Smarter Services]*, Vol.38, No.4, 2011, pp. 87-93.

21. Zheng, Y., Liu, L., Wang, L., Xie, X., Learning Transportation Mode from Raw Gps Data for Geographic Applications on the Web, *Proceedings of the 17th International Conference on World Wide Web*, 2008, pp. 247-256.